\documentclass[sigconf]{acmart}

\renewcommand\footnotetextcopyrightpermission[1]{} % removes footnote with conference information in first column
\usepackage{ulem}
\usepackage{tcolorbox}
\usepackage{xspace}
\usepackage{enumitem}

\def\BibTeX{{\rm B\kern-.05em{\sc i\kern-.025em b}\kern-.08emT\kern-.1667em\lower.7ex\hbox{E}\kern-.125emX}}
    
\copyrightyear{2019}
\acmYear{2019}
\setcopyright{acmlicensed}

\newcommand{\etal}[0]{\textit{et al.}\xspace}

\begin{document}

%
% The "title" command has an optional parameter, allowing the author to define a "short title" to be used in page headers.
\title{Modeling Vocabulary for Source Code Language Models}
\title{Modeling Vocabulary for Large-Scale Source Code Machine Learning}
\title{Modeling Vocabulary for Big Code Machine Learning}
% \title{modeling Source Code Vocabulary for Neural Machine Learning}
%\title{modeling Source Code Vocabulary}

%
% The "author" command and its associated commands are used to define the authors and their affiliations.
% Of note is the shared affiliation of the first two authors, and the "authornote" and "authornotemark" commands
% used to denote shared contribution to the research.
\author{Hlib Babii}
\affiliation{%
  \institution{Free University of Bozen-Bolzano}
  \city{Bolzano}
  \country{Italy}
}
\email{hlibbabii@gmail.com}

\author{Andrea Janes}
\affiliation{%
  \institution{Free University of Bozen-Bolzano}
  \city{Bolzano}
  \country{Italy}
}
\email{ajanes@unibz.it}

\author{Romain Robbes}
\affiliation{%
  \institution{Free University of Bozen-Bolzano}
  \city{Bolzano}
  \country{Italy}
}
\email{rrobbes@unibz.it}

%
% By default, the full list of authors will be used in the page headers. Often, this list is too long, and will overlap
% other information printed in the page headers. This command allows the author to define a more concise list
% of authors' names for this purpose.
\renewcommand{\shortauthors}{Babii et al.}

\newcommand*{\numproj}{14,436\xspace}
\newcommand*{\numprojtrain}{10,106\xspace}
\newcommand*{\numprojtest}{2,165\xspace}
\newcommand*{\numprojval}{2,165\xspace}

%
% The abstract is a short summary of the work to be presented in the article.
\begin{abstract}
When building machine learning models that operate on source code, several decisions have to be made to model source-code vocabulary. These decisions can have a large impact: some can lead to not being able to train models at all, others significantly affect performance, particularly for Neural Language Models. Yet, these decisions are not often fully described. This paper lists important modeling choices for source code vocabulary, and explores their impact on the resulting vocabulary on a large-scale corpus of \numproj projects. We show that a subset of decisions have decisive characteristics, allowing to train accurate Neural Language Models quickly on a large corpus of \numprojtrain projects. 
\end{abstract}

%
% The code below is generated by the tool at http://dl.acm.org/ccs.cfm.
% Please copy and paste the code instead of the example below.
%



% This command processes the author and affiliation and title information and builds
% the first part of the formatted document.
\maketitle

\section{Introduction}
Many works have taken advantage of the "naturalness" of software \cite{hindle2012naturalness} to assist a variety of software engineering tasks, including code completion \cite{raychev2014code}, correcting syntax errors \cite{santos2018syntax}, detecting possibly buggy code \cite{ray2016naturalness}, or API migration \cite{phan2017statistical}, among many others \cite{allamanis2018survey}. These approaches analyze large amounts of source code (e.g., hundreds to thousands of software projects), which allows them to build predictive models of various source code properties, using a variety of probabilistic or machine learning (ML) models, inspired by Natural Language Processing (NLP) techniques.

However, the use of NLP techniques in this context---even when taking advantage of the increased structure of software---, relies on the textual nature of source code, and as such rely on a notion of vocabulary. Thus, a crucial early decision to make when modeling source code for a ML task is how to model software's vocabulary. This is all the more important because, unlike in natural language, \textbf{software developers are free to create any identifiers they like, and can make them arbitrarily complex}. This introduces the issue that any model that is trained on a large-scale software corpus has to deal with an extremely large vocabulary. Hellendoorn and Devanbu observe this issue first-hand for the task of Language Modeling, showing that a Neural Language Model (NLM) has difficulties scaling beyond as few as a hundred projects, while a more traditional N-gram language model does not have such an issue \cite{hellendoorn2017deep}. Section \ref{sec:background} details how vocabulary issues impact Language Modeling. 

Given that NLMs (and Neural approaches in general) are the state-of-the-art in the field of NLP, finding ways to scale them to a larger software corpus is very desirable. An additional reason to scale them is that recent results in NLP \cite{howard2018universal, peters2018deep, devlin2018bert} show that NLMs can be used as upstream tasks in a transfer learning scenario, leading to state-of-the-art improvement in downstream tasks.

Section \ref{sec:choices} presents our \emph{first contribution}: a detailed explanation of the possible modeling choices for source code that we identified.
A variety of modeling choices for vocabulary are available, including: which source code elements to include or exclude; whether to filter out unfrequent tokens or not; how to handle different natural languages; and how to handle compound tokens. Some of these choices may have a large impact on vocabulary size, directly impacting the feasibility of training neural approaches.

After listing the possible modeling choices for software vocabulary, we present our \emph{second contribution}: an empirical study of the impact of the modeling choices in practice. Section \ref{sec:vocab} investigate how the modeling choices affect vocabulary size, number of tokens, and out-of-vocabulary (OOV) rate on a large-scale corpus of \numproj projects. We find that the choices have a drastic impact on these metrics, leading to variations in vocabulary of up to \emph{three orders of magnitude}. Importantly, we find that the most common ways to reduce vocabulary (such as splitting identifiers according to case information), are not enough to obtain a vocabulary of a manageable size; advanced approaches such as adaptations of the Byte-Pair Encoding algorithm \cite{gage1994new,sennrich2015neural} are needed to reach this goal.

Following the vocabulary study, we evaluate how these modeling choices impact the training and performance of NLMs. Section \ref{sec:LM} presents our \emph{third contribution}: we find that, with the right set of choices, it is possible to scale NLMs to a large source code corpus: we successfully train several NLMs on a corpus that contains more than \numprojtrain projects. We evaluate two scenarios: language modeling and code completion. Our results show that our language models are competitive with previous approaches, even at very large scales.

We discuss the modeling choices in light of the results (including the implications) in Section \ref{sec:discussion}. Finally, we discuss the limitations of the study in Section \ref{sec:limitations} and close the paper in Section \ref{sec:conclusions}.

%\begin{itemize}
%    \item We document a series of modeling choices for software vocabulary across a variety of dimensions, providing several novel choices. These modeling choices cover several options along six dimensions (Section 4 presents them).
%    \item  We exhaustively evaluate their impact on vocabulary on a large corpus We perform an exhaustive empirical study of these sets of choices on a large software corpus (Allamanis), that contains close to 15,000 software projects; we report on statistics of the resulting vocabularies and corpora in Section 5, finding that some of these set of choices are able to deal with the ``vocabulary explosion'' problem.
%    \item We evaluate the impact of a subset of tasks on the performance for language modeling, a versatile task. To assess how these choices affect actual machine learning tasks, we compare a selection of the sets of choices for language modeling tasks (Section 6); we find that these sets of choices enable us to train neural networks for language modeling even on very large corpora. 
%    \item  We discuss the choices and provide guidelines and guidance. We then discuss the modeling choices in light of our results, providing guidance and guidelines (Section 7). 
 %   \end{itemize}

  %  We close the paper by discussing the limitations of this study (Section 8), before concluding (Section 9).

\section{Background and Related Work}
\label{sec:background}

\subsection{Language Modeling}

A Language Model (LM) estimates the probabilities of sequences of words based on a training corpus. In NLP, these models are used in tasks such as speech recognition \cite{creutz2007morph} or machine translation \cite{jean2014using}. 

\textit{N-gram Language Models.} Traditional language models are based on n-grams: the probability of a token is computed based on the $n$ previous tokens in the sequence. N-gram LMs have shown extensive success in NLP applications. However, n-gram models have two issues. First, they operate on small ranges (the $n$ previous tokens), with usually low values of $n$ (usually 3 to 5; 6 for Java \cite{hellendoorn2017deep}). Increasing $n$ does not scale well if the vocabulary is large: for a vocabulary of size $m$, there are $m^n$ possible n-grams. Second, they suffer from data sparsity: not all possible n-grams are present in the corpus. Smoothing techniques \cite{chen1999empirical} alleviate---but not eliminate---the issue.

\textit{Neural Language Models.} The state-of-the-art in NLP is made of Neural Language Models (NLM) \cite{bengio2003neural}. NLMs represent words in a continuous vector space, which has attractive properties. In these models, words that are semantically similar are close in vector space \cite{mikolov2013distributed}, allowing the model to infer relationships between words, even if they do not appear in a specific context during training. This allows these models to better deal with data sparsity. In addition, some neural architectures such as Recurrent Neural Networks (RNN) \cite{mikolov2010recurrent}, Long Short-Term Memory (LSTM) \cite{hochreiter1997long, sundermeyer2012lstm}, or Transformer \cite{vaswani2017attention} are able to model much longer range dependencies: a study of LSTM language models showed that they use context as far as 250 words \cite{khandelwal2018sharp}. In addition, NLMs have shown to be versatile: recent work shows that NLMs can be used as upstream tasks for transfer learning. The intuition behind this is that a model that is trained to predict the next word given a sequence of words has to learn features that are useful for other tasks, such as syntax \cite{blevins2018deep}. This property is very attractive since language modeling is an unsupervised task, while the downstream tasks are often supervised tasks. An LSTM LM can be re-purposed for classification tasks by replacing its last layers (performing word predictions) with layers performing classification, before fine-tuning it on the new task \cite{howard2018universal}. Similar results have been shown for a variety of additional NLP tasks, including question answering and entailment \cite{peters2018deep}, sequence to sequence models for translation or summarization tasks \cite{ramachandran2016unsupervised}. Unidirectional \cite{radford2018improving} and bidirectional \cite{devlin2018bert} Transformer models can also be fine-tuned for a variety of downstream tasks, while even larger models show adaptation to downstream tasks with no or extremely little fine-tuning \cite{radford2019language}. Many of these tasks showed state-of-the-art improvements stemming from this pre-training.

% Ramachandran \etal found that initializing a sequence-to-sequence models with two language models yielded improvements in translation and summarization tasks \cite{ramachandran2016unsupervised}. Peters \etal found that word embeddings extracted from the internal states of a language model improve performance in a variety of scenarios including question answering and entailment \cite{peters2018deep}. Howard and Ruder show how to pretrain an LSTM language model and fine-tune it for supervised classification tasks \cite{howard2018universal}. Radford \etal train a Transformer language model and fine-tune it for 12 target tasks \cite{radford2018improving}. Devlin \etal show further improvements by training a bidirectional Transformer model \cite{devlin2018bert}.  Finally, Radford \etal show that a very large Transformer language model, trained on 40GB of text scraped from web pages, it is able to perform a variety of tasks without fine-tuning \cite{radford2019language}.

\textit{Language Models in Software Engineering.} Seminal studies have laid the groundwork for the use of language models on source code: Gabel and Su show that software is very repetitive \cite{gabel2010study}. Hindle \etal compare software to natural language, finding that software is much more repetitive than natural language \cite{hindle2012naturalness}; they build language models of source code, finding applications in code completion. Tu \etal \cite{tu2014localness} find that software is even more repetitive taking local context into account. Rahman \etal refines those results and finds that while some aspects of software are not as repetitive as previously thought (non-syntax elements), others are even more so (API sequences) \cite{rahman2019revisiting}. Allamanis \etal describe the field of probabilistic models of source code \cite{allamanis2018survey}; we cover a subset of these works below.

%\vspace{-12}
\subsection{Large Vocabularies in Machine Learning}
ML models in general, and Language Models in particular, do not deal well with large vocabularies. Since most ML algorithms work on numerical data, text has to be converted to a numerical representation. As part of pre-processing, words are converted to vector representations via one-hot-encoding, producing (sparse) vectors of length equal to the vocabulary. NLMs convert these to word embeddings, dense word vectors of much smaller dimensions (usually in the hundreds), in their first layer. Given enough training data, words that are close in this vector space are semantically similar, and some arithmetic operations are semantically meaningful (e.g., the closest vector to the sum of \texttt{"Germany"} and \texttt{"capital"} is the vector corresponding to \texttt{"Berlin"} \cite{mikolov2013distributed}). For a vocabulary of size $m$ and embeddings of size $n$, the embedding layer is represented by a dense matrix of size $m \times n$.

This solution is not without issues: first, the vocabulary must be known in advance and will be built based on the training corpus. Any new word will not be able to be one-hot encoded as the resulting vector would exceed the expected dimensions. (Some NLM fine-tuning approaches allow the addition of new words by adding rows in the embedding matrix; those are initialized with the mean of the embedding matrix.) A common workaround is to have a specific \emph{unknown} token, and replace any word not previously seen by this token. This is far from ideal, as this amounts to losing information. %\rr{add more on OOV and open vocab (plus LM fine-tuning there)}. 
Second, embeddings are usually computed based on word co-occurrences; deriving meaningful embeddings for rare words is difficult since there is very little data to work with. Third, approaches that generate sequences, such as language models or translation approaches, must output probability distributions that span the entire vocabulary. This is usually done with a Softmax layer; unfortunately, this operation scales linearly with the size of the vocabulary.  For an efficient language model implementation, most of the computation can be spent in the Softmax layer for even a small vocabulary of 10,000 words \cite{bradbury2016quasi}. For larger vocabularies, it can dominate computations; Jozefowicz \etal qualify a Softmax operating on a vocabulary of 800,000 elements as ``prohibitively slow'' \cite{jozefowicz2016exploring}. Further, vocabulary size impacts the size of the model as both the embedding layer and the Softmax layer depend on it. This increases memory requirements for the model, which might impact other parameters (e.g., decreasing batch size to have the model fit in memory, slowing down training; or decreasing Back Propagation Through Time, thus shrinking available context). Finally, a very large vocabulary negatively impacts performance (particularly when out-of-vocabulary words are present \cite{jean2014using}).

While the Softmax issue is specific to the Language Modeling task, \emph{any Neural Model operating on textual sequences will be affected by the other issues}. Further, Neural Models using more complex structures (e.g., trees \cite{alon2019code2vec} or graphs \cite{allamanis2017learning}) also need to model the textual aspects of source code.

%\vspace{-12}
\subsection{Handling Large Vocabularies}
Several approaches exist to deal with large vocabularies.

\textit{Softmax improvements.} The Softmax operation can become a bottleneck even for low vocabularies \cite{bradbury2016quasi}. Several approaches have been proposed to make the computation of the Softmax more efficient. Goodman proposes to speed up maximum entropy language models by grouping words (via clustering) in classes, and divide the operation in predicting a class, before predicting a word among a subset \cite{goodman2001classes}. Morin and Bengio adapt this approach to Neural LMs \cite{morin2005hierarchical}. Bengio and Sen\'ecal propose the use of importance sampling to train a Neural LM \cite{bengio2003quick}, although this approach provides a speedup only during training, not inference. Importance sampling was later applied in Neural Machine Translation for large vocabularies (500,000 words) \cite{jean2014using}. Grave \etal propose an adaptation of the hierarchical Softmax that is efficiently computed on a GPU \cite{grave2017efficient}.

\textit{Approaches with subwords.} Another set of approaches focuses on the issues of vocabulary size, and modeling rare or out-of-vocabulary words with subwords. Creutz \etal observe that ``morphologically rich'' natural languages such as Finnish, Estonian, and Turkish pose issues for language models as their vocabulary can be very large \cite{creutz2007morph}. They decompose words into subword units called morphemes to build subword n-gram LMs, leading to improvements in speech recognition. Mikolov \etal compare language models at the character level and the subword level (modeling out-of-vocabulary words as sequences of two or three characters), finding that subword models improved on character models \cite{mikolov2012subword}. Sennrich \etal adapt the Byte-Pair Encoding (BPE) algorithm to decompose words into subwords, finding improvements in Neural Machine Translation \cite{sennrich2015neural}. Bojanowski \etal propose to represent words as bags of characters n-grams to compute more descriptive word embeddings, allowing the computation of word vectors for out-of-vocabulary words \cite{bojanowski2017enriching}. Kim \etal combine a character-level convolutional neural network with a NLM \cite{kim2016character}. Vania and Lopez compare various subword decompositions (words, morphs, character n-grams, BPE) on several natural languages \cite{vania2017characters}.

\textit{Large Vocabularies in Software Engineering.}
While Hindle \etal observe that in general, source code is more repetitive than natural language \cite{hindle2012naturalness}, Hellendoorn and Devanbu notice that NLMs trained on a software corpus would struggle due to vocabulary size \cite{hellendoorn2017deep}. To produce a model that can be trained in a reasonable amount of time, Hellendoorn and Devanbu impose drastic limits: the number of projects to train on is set to 107 (1\% of the original corpus \cite{allamanis2013mining}), and furthermore, they replace words which occur less than 5 times in the corpus with the ``unknown'' token. Despite this, the resulting vocabulary size is still rather large, totalling more than 76,000 words. Moreover, the prediction performance of a NLM is significantly hurt when it has to predict words that are out of its vocabulary. In parallel to this work, Karampatsis and Sutton \cite{karampatsis2019maybe} also investigate the problem of large vocabularies in source code. Our works are complementary: while we study the impact of various vocabulary choices in depth before training a selection of language models, their work starts with the application of Byte-Pair Encoding and explores language model training in more depth than we do.

%\vspace{-12}
\subsection{Related work} 
Several researchers have developed and exploited probabilistic models of source code; Allamanis \etal \cite{allamanis2018survey} give an overview of this research in a survey. We briefly illustrate the related work dividing it into four parts: constructing language models in software engineering, tackling naming problems, translation approaches, and approaches that aim to model the structure of software systems.

\textit{Constructing Language Models in Software Engineering.} 
Allamanis \etal \cite{allamanis2014learning} develop a framework that learns the style of a codebase to suggests revisions for stylistic consistency. Nguyen \etal \cite{nguyen2013statistical} develop a statistical semantic language model for source code to incorporate semantic information into code tokens and to model the regularities/patterns of such semantic annotations.  Raychev \etal \cite{raychev2014code} address the problem of synthesizing code completions for programs using APIs. They then learn a probabilistic model from existing data and use it to predict properties (e.g., variable names or type annotations) of unseen programs  \etal \cite{raychev2015predicting} . Tu \etal \cite{tu2014localness} introduce a cache language model that consists of an n-gram and an added "cache" component to exploit local regularities. White \etal \cite{white2015toward} motivate deep learning for software language modeling and apply it to code suggestions. Hellendoorn and Devanbu \cite{hellendoorn2017deep} adapt N-gram models for source code (with deeply nested scopes and changing vocabularies) to create language models with a prediction accuracy that surpasses RNNs and LSTM based deep-learning models. %Hindle \etal \cite{hindle2012naturalness} confirm that code can be modeled by statistical languagemodels. 
Nguyen \etal \cite{nguyen2018deep} present a Deep Neural Network language model that complements the local context of lexical code elements with both syntactic and type contexts. Efstathiou \etal \cite{efstathiou2018word} release SE-specific word embeddings trained over 15GB of textual data from Stack Overflow posts, they show the model disambiguates polysemous words better thanks to its SE context. Santos \etal\cite{santos2018syntax} use language models trained on correct source code to find syntax errors, and compare n-gram and LSTM LMs.

\textit{Naming.} Several works predict a name for a source code entity given its context.
Allamanis \etal \cite{allamanis2015suggesting} suggest class and method names with a neural probabilistic language model for source code. They later apply a convolutional neural network with attention to do a similar task \cite{allamanis2016convolutional}. %summarize source code snippets into short, descriptive function name-like summaries using an attentional neural network that employs convolution on the input tokens to detect local time-invariant and long-range topical attention features in a context-dependent way.
Vasilescu \etal \cite{vasilescu2017recovering} describe an approach to recover original names from minified JavaScript programs based on statistical
machine translation (SMT). Bavishi \etal \cite{bavishi2018context2name} accomplish this using a deep learning-based technique.  Jaffe \etal \cite{jaffe2018meaningful} generate meaningful variable names for decompiled code  by combining a translation model trained on a parallel corpus with a language model trained on unmodified C code. 

\textit{Translation approaches.} 
Gu \etal \cite{gu2016deep} propose a deep learning based approach to generate API usage sequences for a given natural language query.  They then propose to learn joint semantic representations of bilingual API call sequences from big source code data to support API call migration  \cite{gu2017deepam}.
Phan \etal \cite{phan2017statistical} use a word2vec model to generate a sequence of C\# API elements and related control units that are needed to migrate a given Java code fragment. Yin \etal \cite{yin2018learning} mine pairs of natural language and code from Stack Overflow to support tasks like code synthesis from natural language. Alon \etal \cite{alon2018code2seq} present an approach that represents a code snippet as the set of compositional paths in its abstract syntax tree and uses attention to select the relevant paths while decoding to generate natural language sequences from source code snippets. Hu \etal \cite{hu2018deep} propose to use NLP and deep neural networks to automatically generate code comments. Tufano \etal \cite{tufano2019learning} investigate the ability of a Neural Machine Translation model to automatically apply code changes implemented by developers during pull requests. 

\textit{Structured data beyond sequences.} Several approaches integrate aspects of the structure of software systems. Note that in each of these cases, vocabulary still needs to be modeled. Allamanis \etal \cite{allamanis2017learning} present how to construct graphs from source code and how to scale Gated Graph Neural Networks training to large graphs. Alon \etal \cite{alon2019code2vec} represent a code snippet as a single fixed-length code vector, which can be used to predict semantic properties of the snippet. Tufano \etal \cite{tufano2018deep} apply deep learning to learn code similarities from different representations. Alon et al. \cite{alon2018general} present a general AST path-based representation for learning from programs to predict program properties such as names or expression types. Ben-Nun \etal \cite{ben2018neural} define an embedding space, inst2vec, based on an Intermediate Representation of the code to recover the semantics of statements based on their context. % Bielik \etal \cite{bielik2016phog} introduce a probabilistic higher order grammar that generalizes probabilistic context free grammars allowing conditioning of a production rule beyond the parent non-terminal, thus capturing rich contexts relevant to programs. 

\newcommand*\choices[0]{\textbf{Choices. }}
\newcommand*\nonEnglish[0]{\texttt{<non-en>}\xspace}
\newcommand*\unk[0]{\texttt{<unk>}\xspace}
\newcommand*\underscore[0]{\texttt{<\_>}\xspace}
\newcommand*\Uppercase[0]{\texttt{<Upper>}\xspace}
\newcommand*\UPPERCASE[0]{\texttt{<UPPER>}\xspace}

\section{modeling choices}
\label{sec:choices}

We present a series of modeling choices for source code vocabulary. These choices may be implicitly made by researchers, without evaluating the alternatives, and may not always be documented in their studies. By making them explicit, we hope that researchers will consider them and document them. Moreover, making them explicit allows us to study their impact on the training and performance of a language models. We make some assumptions behind the modeling choices explicit. The choices that we explore are geared towards specific desirable properties: 

\begin{itemize}[leftmargin=*]
    \item The models should be able to scale to large sizes (thousands of software projects). The training time should not increase much more than linearly as more data is added to the model.
    \item To be versatile, models should avoid losing information. A model should be able to represent the original input as much as possible. Aggressive techniques restrict vocabulary drastically (e.g. to a few hundred tokens \cite{tufano2019learning}), but they lose much information.
    \item Out-of-vocabulary tokens are not desirable, since they prevent a model to reconstruct these tokens. %\rr{This applies to both the training set and the test set.}
    \item Depending on the task, some categories of source code elements (e.g. comments) may not be needed. 
  
\end{itemize}

\subsection{Filtering Infrequent Tokens}

The most common technique is to filter out uncommon tokens (less frequent than a threshold $k$). They are replaced by an \unk token. The advantage of this technique is that it is extremely simple to implement in practice. This modeling choice however loses extensive amounts of information; as such, our goal is to avoid it. % and, if applied indiscriminately, it is \rr{hard to gauge its impact}. 

 %However, the most common workaround, filtering out infrequent words, also conflicts with the objectives. Thus we need to explore further, across the following dimensions: 1) whether non-English source code and comments are necessary; 2) whether string literals and source code comments should be modeled; 3) the role of white space; 4) whether and how to split source code identifiers and literals in subtokens; and 5) if identifiers are split, how to reconstruct them accurately.

\subsection{Natural Language}

Developers may comment their source code in another language, use identifiers in another language, or include non-English literals for testing or internationalization purposes. Thus, source code can contain non-English words in identifiers, strings, and comments. %Needless to say, if a large monolingual corpus already has large vocabulary issues (e.g. the One Billion Words corpus has close to 800,000 words \cite{chelba2013one}), the issue becomes intractable in a multilingual corpus. 
As handling multilingual corpora is an NLP research topic in itself, we evaluate the simplifying assumption to limit a corpus to English. %This must however be done carefully, since many legitimate identifiers would not be found in, e.g., an English dictionary.

We adopt a conservative heuristic to determine that a word is non-English: a word is non-English if it contains non-ASCII characters (we tried dictionary-based heuristics, but they had too many false positives).  This heuristic still has some false positives: words such as ``caf\'e'' and ``na\"ive'' are considered non-English. Non-English words are replaced with a \nonEnglish placeholder. 

This processing filters out non-English words when most of the file is in English, or when the code is in English, but comments and literals may not be. We handle projects that are mostly in another language, and testing/localization files separately. These files would be either non-informative (full of \nonEnglish), or could dramatically expand the vocabulary. During pre-processing, we filter out files with more than a threshold 0.6\% of non-English words in code, or more than 1.9\% in code and strings. The threshold was set by inspecting a sample of files with non-English words, around this threshold, and aiming for no false positives as we favor a conservative approach. This processing removes 0.62\% of files in the corpus; many of them are concentrated in non-English projects that are entirely removed, with the remainder being localization or testing files.

\choices We consider the following choices:
\begin{itemize}[leftmargin=*]
    \item Keep non-English words regardless
    \item Replace non-English words with \nonEnglish. If a token is split (see below), only non-English subtokens will be replaced.
    \item In addition to the previous choice, attempt to remove files that contain a large number of non-English tokens.
\end{itemize}

% Rationale: NLP is still largely monolingual (barring translation). It seems like a reasonable simplifying assumption to remove non-English words.

\subsection{Literals}
In a programming language, tokens have clearly defined types. Some token types have more importance than others: for instance, some types of literals may be less important for some tasks. %\rr{arguably, any literal of importance should be in a variable, the variable being reused}. 

\choices We consider the following choices:

\begin{itemize}[leftmargin=*]
    \item Do not filter any literals.
    \item Have a different minimum frequency for each token type. For instance, the minimum frequency could be higher for numbers, and lower for source code identifiers.
    \item Replace all literals of a specific type with placeholders.
    \item For numbers: only keep ``likely frequent'' numbers, such as numbers less than 100, replacing others with placeholders.
\end{itemize}

% Note that we come back to the case of strings later.

\subsection{Comments and Strings}
Comments and literal strings often contain natural language, rather than source code. Since source code is more repetitive than natural language, one would expect that the contents of comments and string would be much less repetitive. Moreover, while some tasks (e.g. detecting self-admitted technical debt \cite{da2017using}) rely on source code comments, others (such as autocompletion) do not. 

\choices We consider the following choices:
\begin{itemize}[leftmargin=*]
    \item Keep string literals and source code comments intact. Each string literal or comment is modeled as a single token. This choice leads to an explosion of possible tokens, as it models entire sentences or paragraphs as unique tokens.
    \item Keep string literals and source code comments, but model them as sequences of sub-tokens separated by whitespace. This treats these entities as the sequence of words they likely are, and allows to keep all the information in the source code.
    \item If the loss of comments is acceptable, replace comments with \texttt{<comment>} placeholder. Strings are processed as above. 
    \item If the loss of strings is acceptable as well, replace both comments and strings with placeholders (\texttt{<comment>} and \texttt{<string>}).
\end{itemize}

% \rr{Strings and comments are more likely to have non-English content than identifiers}

\subsection{Whitespace}
Some applications (e.g., pretty-printers \cite{allamanis2014learning}) may care about the layout of source code. Others may not, giving importance only to syntactic or semantic aspects (unless code layout is syntactically important, such as in Python). Note that whitespace has a negligible effect on vocabulary, as less than a handful of distinct tokens are needed. It does however significantly increase the amount of tokens in the final corpus.

\choices We consider the following choices:
\begin{itemize}[leftmargin=*]
    \item Tabs, spaces, and newlines are modeled as an individual token.
    \item Different tokens are used to represent two tabs, three tabs, etc.
    \item Formatting is not important: tabs and newlines are removed.
\end{itemize}

\subsection{Word Splitting and Casing}

\textit{Word splitting.} At 70\% of source code \cite{deissenboeck2006concise}, identifiers are the bulk source code and its vocabulary. %particularly as identifiers can be created at will by developers. 
While new identifiers can be created at will, developers tend to follow conventions when creating them. When an identifier is made of several words, it is nearly universal they are visually separated to ease reading, either in \texttt{camelCase} or in \texttt{snake\_case} \cite{binkley2009camelcase}. Thus, an effective way to reduce vocabulary is to \emph{split} compound words according to these word delimiters.

\textit{To split, or not to split.} The decision whether to split compound words or not has important ramifications. First, it introduces additional complexity: the LM can no longer rely on the assumption that source code is a sequence of tokens. Instead, compound words are modeled as sequences of subtokens. Predicting a compound word in a large vocabulary becomes predicting a sequence of subtokens, albeit in a smaller vocabulary. Note that in some cases (e.g., machine translation), techniques such as beam search can be used to keep track of more than one prediction. Second, subtokens increases the length of the sequences, making it harder to relate the current subtokens to the past context, as it increases in size. 

On the other hand, splitting tokens has advantages: the most obvious one is that the vocabulary can be---drastically---smaller. The second is that the out-of-vocabulary rate can be significantly reduced as a consequence. A third is that the model may be able to infer relationships  (e.g. via embedding) between subwords, even if the composed word is rare, as the subwords are more common than the composed word. Approaches using subtokens have shown that splitting tokens allows a model to suggest \emph{neologisms}, tokens unseen in the training data \cite{allamanis2015suggesting}. 

%\rr{some related work on identifier splitting and abbreviation expansion: Guerrouj on normalizing source code, studies on splitting in feature location ... }

\textit{Word casing.} A subsequent decision is whether and how to keep case information. By default, words in different case (e.g. \texttt{value}, \texttt{Value}, \texttt{VALUE}) will be distinct words for the LM. This could cause the vocabulary to increase by a factor of up to 3 times, and make it harder for the LM to infer that words are similar. On the other hand, entirely removing case information loses information. Our solution is to encode case information in separator tokens (e.g., \underscore, \Uppercase, \UPPERCASE ), at the cost of further increasing the size of the sequences. Table \ref{tab:seq} provides examples of how we encode compound words. Other encodings could further reduce the number of tokens.

{\small\begin{table}[h] 
\small{
\centering
\caption{Example word splits}
\begin{tabular}{l}
\toprule
Lowercase\\
\midrule
\texttt{<w> <Upper> malformed <UPPER> url <Upper> exception </w>} \\
\texttt{<w> <UPPER> layout \_ <UPPER> inflater \_ <UPPER> service </w>} \\
\texttt{<upper> tokenbreakingconventions} \\
\midrule
Case preserving \\
\midrule
\texttt{<w> Malformed URL Exception </w>}\\
\texttt{<w> LAYOUT \_ INFLATER \_ SERVICE </w>}\\
\texttt{Tokenbreakingconventions}\\
%Total               & 408,448 &      \Chart{1.00} \\
\bottomrule
\end{tabular}
\label{tab:seq}
}
\end{table}}

\choices 
The following decisions are possible:
\begin{itemize}[leftmargin=*]
    \item Keep tokens as is, unsplit.
    \item Split tokens in subtokens, according to case, and keep it.
    \item Split tokens in subtokens, according to case, and encode case in separator tokens.
    % \rr{If the ability to reconstruct source code is important, the underscore should be kept}.
\end{itemize}

%Whether they should be split or not is a key decision. Pros/cons of splitting vs not splitting in general \rr{Beam search for reconstructing them in "Translation" setting?} Splitting introduce the complexity of a new type of sequences. better possibility to related words, particularly rare ones.

\subsection{Subword Splitting}
Even with word splitting, vocabulary may still grow large. First, natural vocabulary \emph{is} large: many similar words (plural forms, past tenses, etc) will be modeled as entirely distinct words. Developers may not follow conventions to separate words, or the conventions may always apply (e.g. package names in Java are in lower case). Finally, identifiers can contain arbitrary numbers or sequences of characters (such as auto-generated identifiers). If we split in subtokens in the first place, why not go even further?

\textit{Character models.} At the extreme, words are sequences of characters. The vocabulary needed would just be the set of possible characters; the out-of-vocabulary issue vanishes. Unfortunately, this drastically inflates sequence lengths, so a character model is not desirable. However there are interesting intermediate choices.

\textit{Numbers.} Numbers are responsible for a large proportion of the vocabulary, yet \emph{their} vocabulary is very limited. Thus, an alternative to filtering them out is to model them as a sequence of digits. 

\textit{Byte-pair encoding.} Byte-Pair Encoding (BPE) is an algorithm originally designed for data compression, in which bytes that are not used in the data replace the most frequently occurring byte pairs or sequences \cite{gage1994new}. This approach has been adapted to build vocabularies in NLP \cite{sennrich2015neural}: the most frequently occurring sequences are merged to form new vocabulary words. The only parameter BPE needs is the number of merges ($n$) to do.  BPE starts by splitting all the words in characters. Then, it finds the most common pair of successive items in the corpus (initially characters, then tokens). This pair is merged in a new token which is added to the vocabulary; all occurrences of the pair are replaced with the new token. The process is repeated $n$ times. 

% BPE finds the most common sequences of characters, and uses them to build frequently occurring subwords. If frequent enough, a long sequence of characters will be represented by a single word in the vocabulary, while a less frequent one will be represented by several shorter subsequences. 

BPE has several advantages. First, like a character model, no word is out-of-vocabulary; unknown words at test time are represented by subsequences. Second, it dynamically adapts to the frequency of the sequences: common subsequences will be merged, infrequent ones will not. Common words will be represented by a single word (eg, \texttt{exception}), while rare ones will be segmented in roots, prefixes and suffixes (as prefixes and suffixes \emph{are common}). This ensures that each sequence is common enough to have useful embeddings. Finally, it allows for a fine-grained control of vocabulary size, by tuning the number of merges BPE does. A larger vocabulary will have more complete words and less sequences, smaller ones will have longer sequences. %BPE allows each application to find the best threshold to use. %We experiment with the following number of merges when building the vocabulary: 1,000, 5,000, 10,000, and 20,000. %Table \rr{ref!} shows how a selection of words is split with a different number of merges. We see that ...

\choices Excluding character models, the choices are whether to apply BPE on split tokens, and the number of merges to apply.

\subsection{Discarded choices}

We considered stemming \cite{willett2006porter} to reduce vocabulary size but decided against since: 1) stemming approaches work in the context of a specific language, while the corpus is multi-lingual to some degree; 2) stemming loses information: it is not always obvious how to recover the original word from its stem; 3) all words would be stemmed, whereas BPE decomposes infrequent words, keeping frequent words intact. Character-based models result in extremely long sequences, even for very common words; subword models outperform them \cite{mikolov2012subword}. We do not consider aggressive vocabulary abstraction approaches (e.g., replacing identifiers with placeholders \cite{tufano2019learning}, as this prevents recovering the original identifiers.

% paragraph on identifier splitting approaches

\section{Vocabulary results}
\label{sec:vocab}

In this section, we present how the vocabulary modeling choices impact vocabulary at scale. We consider the full Allamanis corpus \cite{allamanis2013mining}, \numproj projects, and study:

\begin{itemize}[leftmargin=*]
    \item \textbf{Vocabulary size.} How large is the resulting vocabulary?
    \item \textbf{Number of tokens.} Several approaches split tokens in subtokens. How does the corpus size grows in response to this?
    \item \textbf{Out-of-vocabulary.} We study the impact of replacing rare tokens with \unk. We report the threshold needed to bring vocabulary size 100K (in line with Hellendoorn and Devanbu's 76K \cite{hellendoorn2017deep}), and the resulting percentage of \unk tokens.
    \item \textbf{Number of projects.} How does the vocabulary grow when more projects are considered?

\end{itemize}

We cover the most important configuration, with the first part of the comparisons shown in Table \ref{tab:vocab}. Each combination is compared to a previous configurations as the baseline.
%We can not compare each combination as they are numerous, but we cover the most . We thus attempt to cover as much as possible in a few cases. The first part of the comparisons is shown in Table \ref{tab:vocab}.

\renewcommand*\unk[0] {\texttt{<unk>}\ }

\newlength\MAX  \setlength\MAX{10mm}
\newcommand*\ChartPer[2]{#1\%~\rlap{\textcolor{black!20}{\rule{\MAX}{2ex}}}\rule{#2\MAX}{2ex}}
\newcommand*\Chart[1]{#1~\rlap{\textcolor{black!20}{\rule{\MAX}{2ex}}}\rule{#1\MAX}{2ex}}

\newcommand*\nostring[0] {\sout{\texttt{"str"}}}
\newcommand*\nocomment[0] {\sout{\texttt{/**/}}}
\newcommand*\yesstring[0] {\texttt{"str"}}
\newcommand*\yescomment[0] {\texttt{/**/}}
\newcommand*\yeswhitespace[0] {\texttt{<tab>}}

\newcommand*\nowhitespace[0] { \yescomment \ \  \yesstring}

\newcommand* \splitnum[0] {\texttt{1/2/3}}

{\small\begin{table*}[h] 
\centering
\caption{Corpus statistics (vocabulary, tokens, out-of-vocabulary (OOV))}
\begin{tabular}{l|rl|rl|l|l}
\toprule
Configuration & Vocabulary & Vs baseline & Corpus Tokens & Vs baseline & 100K filter (OOV) & Baseline \\
\midrule
\midrule
 \multicolumn{7}{c}{Unsplit corpus} \\
 \midrule
Unsplit \yeswhitespace       & 11,357,210 &     \Chart{0.98}  & 2,477,820,538 &     \Chart{0.99} &  200-300 ($\approx$ 4\%) & Unfiltered \yeswhitespace \\
Unfiltered \yeswhitespace     & 11,555,212 &     \Chart{1.00}  & 2,448,156,244 &     \Chart{1.00} &  200-300 ($\approx$ 4\%)  & Unfiltered \yeswhitespace  \\
\midrule
Unsplit \nowhitespace       & 11,357,196 &     \Chart{1.00}  & 1,806,747,721 &     \Chart{0.74} &  200-300 ($\approx$ 5.5\%)  & Unsplit  \yeswhitespace \\
\midrule
% Unsplit \yescomment \ \  \nostring       & 10,265,843 &     \Chart{0.90}  & 1,701,637,340 &     \Chart{0.94} &  200-300 ($\approx$ 5.5\%)  & Unsplit \nowhitespace \\
Unsplit \nocomment \ \  \yesstring      & 10,753,203 &     \Chart{0.95}  & 1,238,234,376 &     \Chart{0.69} &  200-300 ($\approx$ 7.5\%)  & Unsplit \nowhitespace \\
Unsplit \nocomment \ \  \nostring & 9,499,013 &     \Chart{0.84}  & 1,133,050,827 &     \Chart{0.63} &  150-200 ($\approx$ 6.5\%)  & Unsplit \nowhitespace \\
%Total               & 408,448 &      \Chart{1.00} \\
\midrule
\midrule

 \multicolumn{7}{c}{Word splitting (compoundWord \textrightarrow \  compound <cap> word) } \\
%Configuration & Vocabulary & Vs baseline & Corpus Tokens & Vs baseline & 100K filter (oov) & Baseline \\
\midrule

Split \nowhitespace  & 1,588,777 &     \Chart{0.14}  & 2,972,812,831 &     \Chart{1.65} &  45-50 ($\approx$ 0.21\%)  & Unsplit \nowhitespace \\
Split \nocomment \ \  \yesstring  & 1,382,189 &     \Chart{0.13}  & 2,245,853,706 &     \Chart{1.81} &  30-35 ($\approx$ 0.21\%)  & Unsplit \nocomment \ \  \yesstring \\
%Split \yescomment \ \  \nostring  & 1,238,330 &     \Chart{0.12}  & 2,814,362,583 &     \Chart{1.65} &  35-40 ($\approx$ 0.17\%)  & Unsplit \yescomment \ \  \nostring \\
Split \nocomment \ \  \nostring  & 974,606 &     \Chart{0.10}  & 2,087,403,458 &     \Chart{1.84} &  20-25 ($\approx$ 0.15\%)  & Unsplit \nocomment \ \  \nostring \\
\midrule
\midrule
 \multicolumn{7}{c}{Spliting numbers (123 \textrightarrow \  1 2 3)} \\
\midrule
Splitnum \nowhitespace  & 999,885 &     \Chart{0.63}  & 3,045,857,316 &     \Chart{1.02} &  35-40 ($\approx$ 0.15\%)  & Split \nowhitespace \\
Splitnum \nocomment \ \  \yesstring & 832,994 &     \Chart{0.60}  & 2,295,315,822 &     \Chart{1.02} &  25-30 ($\approx$ 0.14\%)  & Split \nocomment \ \  \yesstring \\
%Splitnum \yescomment \ \  \nostring  & 724,133 &     \Chart{0.58}  & 2,875,600,408 &     \Chart{1.02} &  25-30 ($\approx$ 0.11\%)  & Split \yescomment \ \  \nostring \\
Splitnum \nocomment \ \  \nostring  & 504,660 &     \Chart{0.52}  & 2,125,058,914 &     \Chart{1.01} &  20-25 ($\approx$ 0.10\%)  & Split \nocomment \ \  \nostring \\
%Total               & 408,448 &      \Chart{1.00} \\
\midrule
\midrule
% Configuration & Vocabulary & Vs baseline & Corpus Tokens & Vs baseline & 100K filter (oov) & Baseline \\
 \multicolumn{7}{c}{ASCII filtering (\"uber \textrightarrow \ <non-English>)} \\
\midrule

ASCII \nowhitespace  & 978,089 &     \Chart{0.98}  & 3,045,857,316 &     \Chart{1.00} &  35-40 ($\approx$ 0.15\%)  & Splitnum \nowhitespace \\
ASCII \nocomment \ \  \yesstring & 817,742 &     \Chart{0.98}  & 2,295,315,822 &     \Chart{1.00} &  25-30 ($\approx$ 0.14\%)  & Splitnum \nocomment\ \  \yesstring \\
% ASCII \yescomment \ \  \nostring  & 714,667 &     \Chart{0.99}  & 2,875,600,408 &     \Chart{1.00} &  20-25 ($\approx$ 0.10\%)  & Splitnum \yescomment \ \  \nostring \\
ASCII \nocomment \ \  \nostring  & 504,431 &     \Chart{1.00}  & 2,125,058,838 &     \Chart{1.00} &  20-25 ($\approx$ 0.15\%)  & Splitnum \nocomment \ \  \nostring \\
\midrule
\midrule
% Configuration & Vocabulary & Vs baseline & Corpus Tokens & Vs baseline & 100K filter (oov) & Baseline \\
 \multicolumn{7}{c}{Keeping case (compoundWord \textrightarrow \  compound Word) } \\
\midrule
Case \nowhitespace  & 1,231,375 &     \Chart{1.26}  & 2,593,099,484 &     \Chart{0.85} &  70-75 ($\approx$ 0.30\%)  & ASCII \nowhitespace \\
Case \nocomment \ \  \yesstring & 900,806 &     \Chart{1.26}  & 2,440,806,776 &     \Chart{0.85} &  55-60 ($\approx$ 0.23\%)  & ASCII \nocomment \ \  \yesstring\\
% Case \yescomment \ \  \nostring  & 1,034,752 &     \Chart{1.27}  & 1,936,173,045 &     \Chart{0.84} &  50-55 ($\approx$ 0.30\%)  & ASCII \yescomment \ \  \nocomment \\
Case \nocomment \ \  \nostring  & 635,517 &     \Chart{1.26}  & 1,783,880,337 &     \Chart{0.84} &  35-40 ($\approx$ 0.20\%)  & ASCII \nocomment \ \  \nostring \\
%Total               & 408,448 &      \Chart{1.00} \\
\bottomrule
\end{tabular}
\label{tab:vocab}
\end{table*}}

\subsection{Unsplit models}

\textit{Full model.} Our most complete configuration is ``Unsplit full'': it contains all the files in the (de-duplicated \cite{allamanis2018adverse}) corpus, including whitespace, comments, literals, and unsplit tokens. The only pre-processing it has is that comments and strings are modeled as sequences of words, rather than whole entities (doing so would roughly double the vocabulary). This vocabulary contains an excess of 11,5 million unique words. %If, similar to Hellendoorn and Devanbu, we take an upper limit of 100,000 as a workable vocabulary size, we exceed this by a factor of 115. 
To reduce the vocabulary to less than 100,000 words involves replacing words that appear less than \emph{200 to 300 times in the corpus}. The \unk token would be 4\% of the corpus and would be \emph{the 6th the most frequent token}.

\textit{Non-English files.} While our heuristic to remove non-English files is conservative, it has little effect: it reduces vocabulary size by roughly 2\% (200,000 tokens), and removes roughly 1\% of the tokens.

\textit{Whitespace.} Models that do not need whitespace can reduce the size of the corpus by 25\% by removing spaces, tabs, and newlines. However, among non-whitespace tokens, the amount of \unk tokens needed to reach a 100K vocabulary is even higher. 

\textit{Replacing comments and strings with placeholders (\nocomment, \nostring).} Both methods reduce vocabulary, by 5\% for comments and a further 11\% for strings. Removing both makes the vocabulary smaller than 10 million words. However the proportion of \unk tokens to reach a 100K vocabulary rises to 6.5\% (the 5th most common token). It appears source code tokens are more varied than comments and strings. Removing comments reduces corpus size by 30\%.

\newcommand*\conclusion[1]{
\begin{tcolorbox}[left=0mm,right=0mm,boxrule=0.25mm,colback=gray!5!white]
\vspace{-0.2cm}
#1
\vspace{-0.2cm}
\end{tcolorbox}
}

\conclusion{Modeling tokens without splitting them in sub-tokens is impossible to do at scale without extremely aggressive filtering. \unk would be one of the most frequent tokens.}

\subsection{Word splitting}

\textit{Word splitting.} %Since keeping entire tokens is clearly not feasible at large scales, we investigate the effectiveness of splitting tokens based on common coding conventions (camelCase and snake\_case). 
The effect of splitting according to camelCase and snake\_case is considerable: the vocabulary reduces by a factor of up 7 to 10, depending on the presence of strings and comments. The decrease is larger for models without strings and comments, who are richer in compound identifiers; the split model reaches a size of less than one million tokens. The flipside is that the number of tokens in the corpus considerably increases, as compound words are now sequences (including tokens encoding case): the corpus increases by 65 to 84\%. To reach 100K words, the thresholds are much lower (even if still high, ranging from 20 to 50). The percentage of tokens that are \unk is also much lower (0.15--0.21\%). 

\textit{Splitting numbers.} While the improvement is important, the vocabulary is still extremely large. Splitting numbers in digits yields a considerable decrease vocabulary from more than a third to nearly half, at the cost of a very modest increase in number of tokens (1--2\%). Thus splitting numbers (or replacing them with placeholders) is very effective: the smallest configuration hovers just above half a million tokens---a 23 times improvement over the initial one.

\textit{Non-English words.} Filtering non-English words by the ASCII encoding heuristic offers very limited improvements: either the heuristic is too conservative, or much of the improvement was already done in the initial filtering of non-English files. (The heuristic is much more effective for BPE.)

\textit{Keeping case.} The benefit of keeping case is that compoound words are described by shorter sequences, as case-encoding tokens are no longer needed. Compared to equivalent configurations, it decreases the size of the corpus by 15\%, but increases vocabulary by 25\%. %Whether this is a valuable tradeoff depends on individual applications. 
Keeping case could have increased vocabulary by anything from 1 time to 3 times; so 1.25 times is in the lower range of estimates.

\conclusion{Word splitting heuristics are very useful to decrease vocabulary size, at the cost of increasing corpus size. However, the vocabulary is, at best, five times more than our pre-defined threshold.}

\subsection{Vocabulary growth}

\begin{figure}[h]
    \centering
    \includegraphics[width=0.45\textwidth]{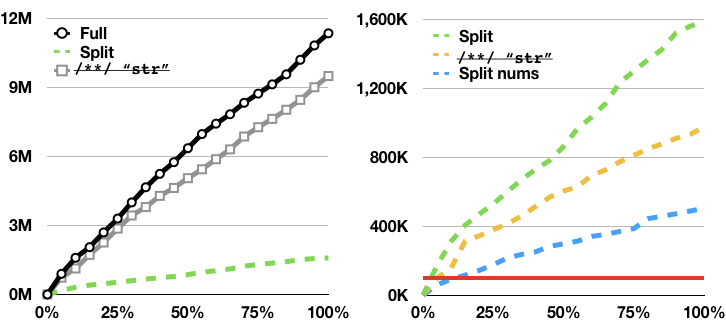}
    \caption{Growth of vocabulary for the corpus}
    \label{fig:growth}
\end{figure}

While these results are encouraging, the growth of the vocabulary as projects are added provides another perspective. Figure \ref{fig:growth}, left compares the growth of the vocabulary size for unsplit configurations, and for the largest of the split corpora. The difference is large and widens significantly as more projects are added.

Figure \ref{fig:growth}, right shows growth of vocabulary for three split configurations. We also see widening gaps between the configurations. However, \emph{none of the curves appear to plateau}. There is no indication that the vocabulary will stabilize at some point. Going from 75 to 100\% of projects with the best configuration adds nearly 20\% new words. \emph{As unseen projects are added, out-of-vocabulary words are more likely}.

% While fine-tuning language models do support adding new words to their vocabulary to some extent (e.g. by initializing their embedding to the mean of the other embeddings \cite{howard2018universal}), this issue, coupled with the large vocabulary size, tells us this is not enough.

\conclusion{Vocabulary grows in an apparent linear fashion as new projects are added; word splitting is not enough to get it under control}

\subsection{Byte-Pair Encoding}

\newcommand{\spl}[0]{\begin{picture}(2,0)(-1,-3)\circle*{3}\end{picture}\ }

BPE allows us to specify our vocabulary size. Thus, the question is not whether it reduces vocabulary---it will!---, but how much is a good tradeoff between vocabulary size and sequence size.

\emph{Quantitative evidence.} Corpus sizes for BPE configurations are shown in Table \ref{tab:vocabbpe}. If we compare the models with our lowest vocabulary (ASCII, numbers split, \nocomment, \nostring), to vocabularies obtained with BPE, we see that a vocabulary of (slightly more) than 1,000 words grows the corpus size by 22\%. Most interestingly, a similar model with 5,000 words (20 times \emph{less} than the 100K threshold), grows it by 4\%. Finally a model with 10,000 words grows the corpus by only 1\%, but reduces the vocabulary by a factor of \emph{50}! (Note that including non-ASCII words add a considerable amount of unicode characters, growing the vocabulary by $\approx$ 5,000 words in each case.)

In addition, since BPE merges based on frequency, the resulting subtokens, no matter their size, are frequent. Depending on the configuration, between 91 and 96\% of subtokens occur more than 1,000 times in the corpus, and 97--99\% occur more than 100 times. The fact that the items are frequent means that it is much more likely that good embeddings can be computed for therm.
%\textbf{In addition, filtering words by frequency barely has an effect on BPE's vocabulary size. This means that \emph{most subwords are very frequent, enabling the computation of better embeddings}.}

{\small\begin{table}[h] 
\small {
\centering
\caption{Vocabulary statistics for BPE variants}
\begin{tabular}{l|rl|l}
\toprule
Configuration & \footnotesize{Tokens (M)} & Vs baseline & Baseline \\
\midrule
1K \nostring \nocomment & 2,600 &     \Chart{1.22} & ASCII \nostring \nocomment \\
5K \nostring \nocomment & 2,209 &     \Chart{1.04} & ASCII \nostring \nocomment \\
10K \nostring \nocomment & 2,153 &     \Chart{1.01} & ASCII \nostring \nocomment \\
\midrule
5K \yesstring\nocomment & 2,396 &     \Chart{1.04} & ASCII \yesstring\nocomment \\
10K \yesstring\nocomment & 2,328 &     \Chart{1.01} & ASCII \yesstring\nocomment \\
Case 5K \yesstring\nocomment & 2,173 &     \Chart{1.12} & Case \yesstring\nocomment \\
Case 10K \yesstring\nocomment & 2,043 &     \Chart{1.06} & Case \yesstring\nocomment \\
\midrule
5K \yesstring\yescomment & 3,228 &     \Chart{1.06} & ASCII \yesstring\yescomment\\
10K \yesstring\yescomment & 3,095 &     \Chart{1.02} & ASCII \yesstring\yescomment\\
Case 10K \yesstring\yescomment & 2,753 &     \Chart{1.06} & Case \yesstring\yescomment\\
Case 20K \yesstring\yescomment & 2,647 &     \Chart{1.02} & Case \yesstring\yescomment\\
\bottomrule
\end{tabular}
\label{tab:vocabbpe}}
\end{table}
}

\emph{Qualitative evidence.} We inspected 110 random identifiers longer than 25 characters long, alongside with the splits produced by BPE---the rationale being that longer identifiers are more likely to provide interesting splits. We show some examples in Table \ref{tab:bpes}. While some of the words have optimal splits even at BPE 1K (example a), some are clearly sub-optimal (example b), but are optimal at BPE 5K. The model handles rare words due to typos gracefully (example c), and splits words correctly without case information, with enough merges (example d). Some words have satisfying splits at low BPEs, yet improve as BPE increases (example e). Finally, the model degrades gracefully for non-English words: those are not out-of-vocabulary, just long sequences (example f).

We classified each of the 110 splits in 3 categories: \emph{good} (reproduces the expected case split), \emph{acceptable} (one word was split in root and prefix or suffix, such as Grid \spl ify, or an acronym was not well reconstructed, such as I \spl BAN), and \emph{degraded} (one or more words split incorrectly, or in more than 2 parts). We found 7 degraded splits: 2 foreign words, 2 words with typos (\texttt{Fragement, INCULDED}), 1 with rare words (\texttt{TheImprisonedGourmet}), an all-lowercase sequence of 8 words, and a word were the split was unclear (\texttt{appirate}). Of the \emph{good} splits, 11 were found at BPE 1K (including common words such as \texttt{exception}, \texttt{configuration}, or \texttt{attribute}), 51 at BPE 5K, 28 at BPE 10K, and 8 at BPE 20K. While BPE 1K is too small, 5K is competive, 10K is optimal, and 20K offers disminishing returns. 

\emph{Adding back string, comments, and case.} Encouraged by these results, we increased the base vocabulary. We find that adding words found in strings and comments appears to have little impact on BPE 5K and 10K, both of which slightly increase the size of the corpus by 1--2\%. A vocabulary of 10K words is more than 1,000 times smaller than the initial configuration (11,357,210), at the cost of increasing the number of tokens in the corpus by a factor of 1.7.

However, adding back case has a larger impact, as a relatively large number of words have at least two versions (example g). A second manual inspection of the same splits revealed that more words were decomposed in subwords (e.g., \texttt{adjusted} becomes \texttt{Adjust}\spl{ed}, or \texttt{implicitly} becomes \texttt{Implicit}\spl\texttt{ly}). Raising the amount of merges to 20,000 is necessary, but it increases the corpus by 2\% only, for a corpus with strings and comments.

We conclude that as a rule of thumb, a BPE with a 1--2\% token increase performs very well. We note that our BPE also includes all numbers and literals: some sequences that were merged were common numbers. An approach that can afford to filter uncommon literals and numbers with a low out-of-vocabulary threshold (e.g., 5), may perform even better in the resulting vocabulary.

\conclusion{BPE shrinks source code vocabulary \emph{very} effectively. Moreover, most of the vocabulary is frequent, improving embeddings.}

{\small\begin{table}[h] 
\centering
\small{ 
\caption{Examples of Byte-Pair Encoding Splits}
\begin{tabular}{l|l}
\toprule
Configuration & Token / BPE Split \\
\midrule
\multicolumn{2}{c}{a) Optimal at BPE 1K}\\
\midrule
Original & LAYOUT\_INFLATER\_SERVICE \\
BPE 1K &  layout \spl inflater \spl service \\
\midrule
\multicolumn{2}{c}{b) Optimal at 5K}\\
\midrule
Original & MalformedURLException \\
BPE 1K &  m \spl al \spl for \spl me \spl d \spl url \spl exception \\
BPE 5, 10, 20K & malformed \spl url \spl exception \\
\midrule
\multicolumn{2}{c}{c) Effect of typos}\\
\midrule
Original & INCULDED\_TEMPLATE \\
BPE 1, 5, 10, 20K & inc \spl ul \spl ded \spl template \\
\midrule
\multicolumn{2}{c}{d) Splitting without case}\\
\midrule
Original & cmd\_reloadquestconfig \\
BPE 1K & c \spl m \spl d \spl re \spl load \spl quest \spl config \\
BPE 5,10,20K & cmd \spl reload \spl quest \spl config \\
\midrule
\multicolumn{2}{c}{e) Continuous improvement}\\
\midrule
Original & httpclientandroidlib \\
BPE 1K & http \spl client \spl android \spl li \spl b \\
BPE 5K & http \spl client \spl android \spl lib \\
BPE 10K & httpclient \spl android \spl lib \\
BPE 20K & httpclientandroidlib \\
\midrule
\multicolumn{2}{c}{f) Handling non-English words} \\
\midrule
Original & vormerkmedienauflister \\
BPE 5K & vor \spl mer \spl k \spl medi \spl en \spl au \spl f \spl list \spl er \\
BPE 20K & vor \spl mer \spl k \spl medi \spl en \spl auf \spl lister \\
\midrule
\multicolumn{2}{c}{g) Impact of preserving case} \\
\midrule
Original & alternativeEndpointsAndQueries \\
BPE 5k & alternative  \spl end \spl points \spl and \spl queries \\
BPE 5k (case) & al \spl tern \spl ative \spl End \spl points \spl And \spl Qu \spl eries \\
BPE 10k (case) & alternative \spl End \spl points \spl And \spl Queries \\
\bottomrule
\end{tabular}
\label{tab:bpes}
}
\end{table}}

\section{Training Language Models}
\label{sec:LM}

In this section, we test whether we can successfully train large language models with our vocabulary choices, reporting on some results considering training time. We also consider the model's performance at the language modeling task. This is especially important in light of recent results in NLP that show that the knowledge learned to be able to do unsupervised language modeling effectively can transfer to supervised tasks (see Sect. \ref{sec:background}). While there is early evidence that pre-training NLMs can be useful in Software Engineering NLP tasks, particularly for small datasets \cite{robbes2019leveraging}, applying the same techniques to source code involves solving vocabulary issues. Finally we also consider the more concrete code completion scenario, where we compare our models on some of the code completion scenarios of Hellendoorn and Devanbu \cite{hellendoorn2017deep}.

\subsection{Methodology}

\textit{Models.}
For the NLMs, we use an AWD-LSTM \cite{merity2017regularizing}, a state-of-the-art implementation of the LSTM, with a variety of strategies that improve its regularization capabilities, such as a version of dropout \cite{srivastava2014dropout} adapted for LSTMs. The hyper-parameters were manually tuned on a fraction of the training set; we report on 4 configurations (BPE 5K and 10K, with and without strings). All LSTMs have an embedding layer of size 300, 650 hidden units, and 3 LSTM layers. We set a learning rate of $1_{e-3}$, a weight decay factor of $1_{e-6}$; use the Adam optimizer with parameters 0.7 and 0.99 \cite{kingma2014adam}.
%a \rr{need to go into specifics?}
For the n-gram models, we use the implementation of Hellendoorn and Devanbu.

\textit{Corpus.}
Since our focus is to test whether NLMs can scale, we reuse the large-scale corpus of Allamanis and Sutton \cite{allamanis2013mining}. We divide the corpus in a training set (\numprojtrain projects), a testing set (\numprojtest projects), and a validation set (\numprojval projects). Recent work by Allamanis \cite{allamanis2018adverse} points out that large-scale code duplication can bias the performance of ML approaches; the models train on source files that they may see during evaluation. We thus use the ``dataset errata'' of Allamanis to remove clone groups in the entire corpus.

\textit{Vocabulary choices.}
Our initial goal was to have a configuration as close as possible to the one of Hellendoorn and Devanbu. Similarly to them, we omit source code comments. One difference with the original setup lies in the treatments of string literals: Hellendoorn and Devanbu keep strings, but replace all strings longer than 15 characters with the empty string; we have both kind of models run without strings instead, plus some LSTM variants with strings. (We assumed all strings were kept, and discovered this undocumented behaviour rather late; omitting strings altogether was the choice that had the fewest ramifications.)

Since best performance for NLMs rely on token splitting, we try 2 BPE configurations, one at 5K and another at 10K. Both vocabularies are built on the training set only. For n-grams, we keep tokens unsplit: splitting them would result in the n-gram model reducing its context window and could thus impact performance. 

\textit{Language Modeling performance.}
Language modeling performance is our primary metric of interest as it opens up possibilities for transfer learning. We report the metrics of entropy for all models. However, the entropy depends on vocabulary size and the number of tokens, which vary accross configurations. The effects of this choice are hard to predict: while a model operating on subwords has less vocabulary words to choose from, it also has to make more predictions. Indeed, one could argue that the subword level is a more accurate reflection of true performance, as the percentage of prediction on syntax tokens (e.g., \texttt{;}, \texttt{(}, \texttt{)}, ..., which are extremely common and ``easy to predict'' \cite{rahman2019revisiting}) is lower.

Mikolov \etal compared disparate models (subword and character models) by converting word-level entropy to character-level entropy: $BPC = (\textit{entropy} \times \textit{n\_tokens})/ \textit{n\_chars}$. This was possible since none of the models were predicting out-of-vocabulary tokens. Similarly, we convert subtoken-level to word-level entropies: $\textit{word\_entropy} = (\textit{subword\_entropy} \times \textit{n\_subtokens}) / \textit{n\_tokens}$.

\textit{Code completion performance.}
Although code competion is not our primary focus, we also investigate this task. We report the mean reciprocal rank metric (MRR), similarly to Hellendoorn and Devanbu. MRR is the mean---over all predictions---of the inverse of the rank of the correct choice in a list of prediction: a correct prediction at rank $k$ is scored $1/k$. Similarly to the entropy metric, this metric is provided at the subword level for the NLMs, so a direct comparison with word-level prediction is not possible.

\textit{Evaluation scenarios.}
Hellendoorn and Devanbu have 3 evaluation scenarios, out of which 2 are suitable for NLMs: the static and the dynamic scenario. The static scenario is a cross-project scenario, in which models never train on the test data. The dynamic scenario allows model to train on test data after having seen it, which according to Hellendoorn and Devanbu advantages NLMs. As our primary focus is the transfer learning potential of NLMs, we focus on the static scenario. 

A dynamic scenario could be interesting for fine-tuning a language model on a new project (similarly to \cite{howard2018universal}), but it would be different from the dynamic scenario, as the NLM would be allowed to see the project multiple times; we reserve this fine-tuning for future work, but provide entropy results for n-gram models. 

\textit{Use of cache.}
For n-gram models, Hellendoorn and Devanbu have several cache settings: plain (no cache), cache, and nested caches (various caching following the package structure of software systems). Similarly to before, we do not focus on caches as they do not improve performance in a transfer learning scenario. We evaluate the n-gram model with a cache (the cache-less n-gram did not exhibit good performance).

\textit{Training Speed.}
We also report some metrics on the training speed of NLMs: time to complete an epoch, number of projects per minute, and number of files per second. This gives us insight on the ability these models have to quickly adapt to new data. As n-gram models do not have performance issues, we do not report these.

\subsection{Results}
All the results are presented in Table \ref{tab:results}. We note that some of the NLMs have not yet fully converged, and may improve further with additional training.

{\small\begin{table*}[h] 
\centering
\caption{Performance statistics for a selection of language models}
\begin{tabular}{l|lll|lll}
\toprule
Configuration & Subtoken Entropy & Token entropy & MRR & Min/Epoch & Projects/min & Files/s \\
\midrule
LSTM ASCII BPE 5K (3 epochs) \nostring \nocomment & 1.82 & 3.59 & 0.80 & 375 & 26.9 & 52.7  \\
LSTM ASCII BPE 10K (1 epoch) \nostring \nocomment & 2.19 & 4.22 & 0.78 & 670 & 15.1 & 29.5 \\

\midrule
LSTM ASCII BPE 5K (1.5 epochs) \yesstring \nocomment & 1.96 & 3.84 & 0.79 & 406 & 24.9 & 48.8 \\
LSTM ASCII BPE 10K (1.5 epochs) \yesstring \nocomment & 2.14 & 4.06 & 0.78 & 709 & 14.2 & 27.9 \\
\midrule
6-gram with cache, Unsplit corpus \nostring \nocomment & -- & 5.33 & 0.59 & -- & -- & -- \\
\bottomrule
\end{tabular}
\label{tab:results}

\end{table*}}

\textit{Training Speed.} Our fastest models can process an entire epoch of data (\numprojtrain projects, 1,187,620 files) in roughly 6 hours. This represents roughly \emph{27 projects a minute}, or more than 50 source code files per second. These metrics were computed on a consumer-grade GPU (Geforce GTX 1080, released in 2016). Since one epoch is so large, the models perform well after one epoch only, even if they can improve with more epochs. Models with strings are slower, but have to predict more tokens; the increase is approximately linear. Likewise, doubling the vocabulary roughly slows training by half.

\textit{Language modeling and completion.} We find that with appropriate modeling choices, NLMs can be competitive with N-gram models: the LSTMs have a significantly lower entropy than the N-gram model (1.82--2.19 bits), even when converting it to token-level entropy (3.58--4.22 bits). This is despite the fact that the N-gram model caches, while the LSTMs \emph{never updates on test data}. In addition, two of the LSTMs predict strings as well. Regarding code completion, while a direct comparison is not possible due to the difference in granularities of the predictions, we find that the MRR of the LSTMs are very competitive as well. 

\textit{Dynamic n-gram models.} We also computed entropies for dynamic n-gram models, finding much better token-level entropies of 3.69 bits (no cache), and 2.86 bits (with cache). While our best LSTM still edges out the former, the latter outperforms out all of our LSTMs. This is expected, since maximizing project-level information is helpful for completion; our LSTMs in the fully static setting ignore it. 

\textit{Further training.} We note that training our LSTMs for longer periods (8 epochs) allowed them to outperform the dynamic model, all of them achieving entropies between 3.32 bits and 3.67 bits.

\textit{Performance of n-grams.} The performance of the cached n-gram model is much lower than reported by Hellendoorn and Devanbu. We used their implementation for lexing and parsing; our only change was to convert all strings to empty strings, which should improve performance. We think that two factors cause the difference. The first is the removal of duplicate files (25\% in this corpus). Allamanis also observed significant performance drops when evaluating models on a de-duplicated corpus \cite{allamanis2018adverse}. The second is the size of the test set, and the number of (unsplit) tokens not encountered in the training set, which is very high: the simple cache is not enough to cover all the cases, as the dynamic scenario results show.

\textit{Additional results}
We experimented with n-gram models on split corpora. The reduced vocabulary did increase the performance, but when accounting for subtokens, the change was minimal. We tried higher order n-grams to compensate for the longer sequences, but saw little improvement with 7 or 8-grams.

For the LSTMs, we experimented with a \emph{neural cache}  \cite{grave2016improving}, which is the neural equivalent of a regular cache model.  The neural cache has a fixed-size window of size $n$, in which the $n$ previous activations of the LSTM's last hidden layer are stored with the next word at that time step, at test time.  Importantly, this cache mechanism is a test-time only addition, and does not require training, unlike other alternatives \cite{merity2016pointer}. We did observe slight improvements, but less than we hoped. This is however not surprising, since the neural cache's main benefit is the prediction of out-of-vocabulary tokens, which is not an issue for our model thanks to BPE.

\section{Discussion}
\label{sec:discussion}

\textit{Vocabulary.}
With BPE 10K, the vocabulary size of an ASCII model shrinks by a factor of \emph{more than 1,000} over the initial configuration, and \emph{removes the out-of-vocabulary issue}. On the other hand, the size of the corpus (in number of tokens), is a little less than double (excluding whitespace). Bradbury \etal \cite{bradbury2016quasi} showed that for a QRNN language model, the Softmax layer can start to dominate computation costs for vocabularies as low as 10,000 words. For a similar scenario, reducing the vocabulary a thousand fold, while increasing double corpus size could lead to an execution that is \emph{several hundred times faster}.
More generally, vocabulary has a very large impact on model training and performance, which is why \emph{vocabulary modelling choices should be clearly documented}.

\textit{Training Speed.}
Our models were trained for less than a day on \numprojtrain projects. While they have not fully converged, they already show excellent performance. Such models are able to process \emph{20 to 50 files per second}. While a model may need to see a given file several times to fully integrate it, this is still impressive. Moreover, LSTM variants such as QRNNs \cite{bradbury2016quasi} can  training faster as they better parallelize on the GPU. This has several practical implications:
\begin{itemize}[leftmargin=*]
    \item The slowdown of training larger models with more capacity may be acceptable, yielding potentially higher performance.
    \item Since the vocabulary does not grow, it is easier to train on even more data as training time scales linearly with data.
    \item Taking a generic language model and fully fine-tuning it on a specific project may not be costly, and may be perhaps measured in minutes, rather than hours.
    \item If training is fast, perhaps having dynamic NLMs reacting to context changes is feasible, similar to Hellendoorn and Devanbu's nested caches.
    % \item If updating a model is not too costly, perhaps having more dynamic models with a notion of context
\end{itemize}

%\textit{Broader implications.}
%\begin{itemize}[leftmargin=*]
%\item Papers should document their choices 
%\item If you do this, use that
%\item If you do this other thing, use that other thing
%\end{itemize}

% \textit{Discussion.}

\section{Limitations of this study}
\label{sec:limitations}

\textit{Peer-review data.} Sharing data at this time is impractical: we work with multiple pre-processed versions of a large corpus, and train large language models. Re-running the scripts is also time-consuming. We will release the data, scripts, and models if the paper is accepted.

\textit{Non-exhaustive choices.} While we wanted to cover as many vocabulary modeling choices as we could, we can not guarantee that the choices are exhaustive.

\textit{Filtering.} Our heuristics to filter non-English files are rather crude: they are conservative, and perhaps better heuristics could lead to further vocabulary reductions. On the other hand, some legitimate English words with latin accents may be filtered out. Techniques to recover the unaccented characters such as character folding might improve the model further.

\textit{Handling Non-English languages.} While non-English source code is uncommon in our corpus, non-English comments are more common. Thus a proper handling of other languages would be welcome. % However multilingual language models are still an active research topics.

\textit{Other tasks.} We focus on language modeling. While there is potential to transfer language modeling to other tasks, this potential still has to be realized in future work. 

\textit{Other architectures.} Neural architectures such as QRNNs \cite{bradbury2016quasi} or Transformers \cite{vaswani2017attention} are more computationally efficient than LSTMS. Investigating them would be welcome. More importantly, investigating architectures that take fully advantage of software's structure, via trees \cite{alon2019code2vec} or graphs \cite{allamanis2017learning, tufano2018deep} would be a much better architectural choice. Finally, caching should still be improved for the code completion task. 

\textit{Improvements for code completion.} Our experiments on the code completion scenario were not the main focus of the paper and as could be expanded. A comparison with n-gram and nested caches would be instructive.  Improving on the cache is possible: an example would be a hybrid cache that ``corrects'' a predicted sequence of subtokens if it does not exist in the project, but a similar one does. Another would be adding beam search, if the slowdown is acceptable. In parallel to our work, Karampatsis and Sutton proposed an NLM using BPE, with beam search \cite{karampatsis2019maybe}.

\textit{Improvements on language modeling.} We did not explore the entire space of modeling choices. In particular, perhaps a good tradeoff exists for filtering uncommon literals that are very likely to be unique (e.g. a string of random characters). This would also increase the quality of BPE splits at the same amount of merges. This may enable us to have acceptable quality for an model keeping case information at less than 20K.

\section{Conclusions and future work}
\label{sec:conclusions}

While software is more repetitive than natural language, software vocabulary is much more diverse as developers can create new identifiers at will. This is a serious hurdle to make NLMs work at scale. In this paper, we showed how modeling choices for source code vocabulary drastically influence the resulting vocabulary, and that the techniques that allow the vocabulary to be kept under control (such as BPE) are not necessarily intuitive. 

We showed how applying a set of modeling choices on a large corpus of \numproj software projects made it possible to reduce the vocabulary by three orders of magnitude, while less than doubling the amount of tokens to consider. Further, such a vocabulary is not affected by the out-of-vocabulary problem. 

As a consequence, we were able to train large-scale NLMs at scale: a model trained on \numprojtrain projects can be trained in less than a day, averaging 27 projects per minute and 50 source code files per second. This LM was competitive at language modeling and code suggestion tasks. Moreover, this kind of performance opens the door to the pretraining of large NLMs for software, and the transfer learning possibilities pretrained NLMs enable.

\begin{acks}
The computational results presented in this paper have been achieved using the Vienna Scientific Cluster (VSC).
\end{acks}

%
% The next two lines define the bibliography style to be used, and the bibliography file.
\bibliographystyle{ACM-Reference-Format}
\bibliography{sample-base}

\end{document}